\documentclass{article}

\PassOptionsToPackage{numbers, compress}{natbib}

\usepackage[preprint]{neurips_2024}

\usepackage[utf8]{inputenc} 
\usepackage[T1]{fontenc}    
\usepackage{hyperref}       
\usepackage{url}            
\usepackage{booktabs}       
\usepackage{amsfonts}       
\usepackage{nicefrac}       
\usepackage{microtype}      
\usepackage{xcolor}         
\usepackage{graphicx}
\usepackage{multirow}
\usepackage{xcolor}

\title{Bailing-TTS: Chinese Dialectal Speech Synthesis Towards Human-like Spontaneous Representation}

\author{\small Xinhan Di, Zihao Chen, Yunming Liang, Junjie Zheng, Yihua Wang, Chaofan Ding \\ AI Lab, Giant Network\\} 
\date{}

\begin{document}

\maketitle

\begin{abstract}
Large-scale text-to-speech (TTS) models have made significant progress recently. However, they still fall short in the generation of Chinese dialectal speech. To address this, we propose Bailing-TTS, a family of large-scale TTS models capable of generating high-quality Chinese dialectal speech. Bailing-TTS serves as a foundation model for Chinese dialectal speech generation. First, continual semi-supervised learning is proposed to facilitate the alignment of text tokens and speech tokens. Second, the Chinese dialectal representation learning is developed using a specific transformer architecture and multi-stage training processes. With the proposed design of novel network architecture and corresponding strategy, Bailing-TTS is able to generate Chinese dialectal speech from text effectively and efficiently. Experiments demonstrate that Bailing-TTS generates Chinese dialectal speech towards human-like spontaneous representation. Readers are encouraged to listen to demos at \url{https://giantailab.github.io/bailingtts_tech_report/index.html}.
\end{abstract}

\section{Introduction}
The goal of text-to-speech (TTS) systems \cite{song2024ella,deng2023prosody,choi2024dddm,zhang2024speechalign,wang2024speechx,yu2023language} is to generate human-like speech from text. With the development of neural network and deep learning, large-scale training corpus of a human-like level \cite{wang2017tacotron,shen2018natural,ren2019fastspeech,tan2024naturalspeech,shen2023naturalspeech} is established and corresponding TTS models is developed. However, such system \cite{chen2022wavlm, le2024voicebox, lovelace2023simple, chen2022resgrad, gao2023e3} is only able to generate voice for non-dialectal speech. And the quality of the generated speech is not satisfactory \cite{zhang2023speak, wang2023neural, li2017deep, jiang2023mega, lajszczak2024base, wang2023lm}. Therefore, in order to generate human-like quality of speech from dialectal text, we propose Bailing-TTS, a family of speech generation models for synthesizing Chinese dialectal speech from text with human-level naturalness. 

Besides, in zero-shot speech synthesis, both the naturalness and the robustness of speaker modeling are challenging \cite{popov2021grad}. In previous studies, large-scale models are developed for speech synthesis\cite{le2024voicebox,borsos2023audiolm,jiang2023mega,borsos2023soundstorm,zeghidour2021soundstream}. Despite the advances from these methods, the generalization capability of these methods is still insufficient for the generation of high-quality dialectal speech. Therefore, we primarily propose Bailing-TTS to generate high quality Chinese dialectal speech from text.

We propose several specific extension techniques that significantly enhance the quality of Chinese dialectal speech generation. First, continual semi-supervised learning framework towards text and speech tokens is developed to facilitate the alignment of these multiple modalities. Second, a multiple-stage Chinese dialectal representation learning framework introduces a specific network architecture and the corresponding learning strategy to improve the quality of the generated Chinese dialectal speech. 

In the evaluation of the proposed Bailing-TTS, both the advantages and disadvantages of the proposed model for the dialectal speech generation is compared with human speakers. The evaluation results represent that Bailing-TTS generates good quality of Chinese dialectal speech.

Finally, both the potential applications and the limitation of Bailing-TTS are discussed. Then, the future work of Bailing-family foundation model is represented including taking multiple modalities as input and producing audio output. For example, it's supposed to produce the generation of music \cite{cideron2024musicrl} from the text and video. Besides the generation of audio such as music or speech, in our next step, the video and audio are generated simultaneously. 

Our key contributions are represented as the following:

\begin{itemize}
\item We introduce Bailing-TTS, a family of foundation models for synthesizing Chinese dialectal speech from text with human-level naturalness, such as spontaneous speech.
\item Both a continual semi-supervised representation strategy and Chinese dialectal representation with a specific mixture-of-expert network architecture are proposed to improve the quality of the generated Chinese dialectal speech. 
\item A variety of hierarchical reinforcement post-learning extension techniques are represented to significantly enhance the quality of Chinese dialectal speech generation.
\end{itemize}

\section{Bailing-TTS}
\begin{figure}
    \centering
    \includegraphics[width=1.0\linewidth]{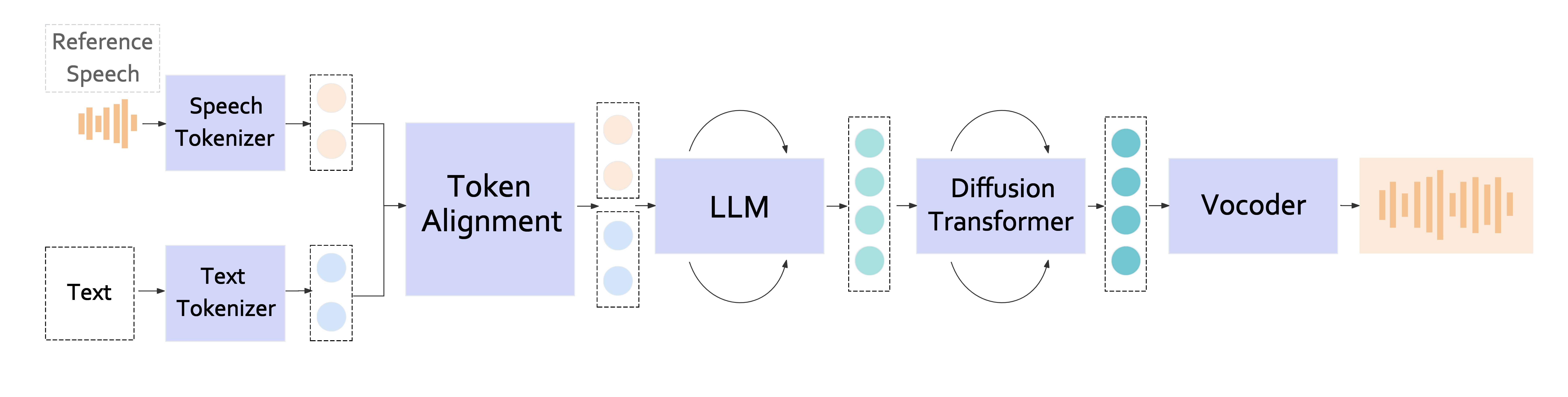}
    \caption{Overall Architecture of Bailing-TTS}
    \label{fig:1}
\end{figure}

We propose Bailing-TTS based on an auto-regressive transformer model of multiple layers. The proposed Bailing-TTS is trained on a large-scale of dataset including a large part of dialectal data. Both non-precise annotation data and high-quality annotation data are in the dataset. Besides, we propose a multi-stage training strategy, based on a specific transformer architecture (Figure \ref{fig:1}), to facilitate the generation of spontaneous and expressive speech from text.

\subsection{Overall Architecture}
First, at the stage of token representation, a continual semi-supervised learning strategy of spontaneous, expressive text and speech token pairs is proposed to facilitate weak alignment between the two modalities. Second, a specific transformer-based architecture is proposed corresponding with a multi-stage training strategy. Third, during inference, it generates good quality of spontaneous Chinese dialectal speech from text. 

\subsection{Continual Semi-supervised Learning Towards Text and Speech Tokens}
Spontaneous speaking style exhibits notable differences from other speaking styles due to various spontaneous phenomena (e.g., filled pauses, prolongation) and substantial prosody variation (e.g., diverse pitch and duration variation), posing challenges to modeling and prediction of spontaneous style. 

In order to generate spontaneous, expressive speech from text, we present a multi-stage and multi-modal (text and speech) pre-training learning framework for text-speech token alignment (Figure \ref{fig:1}). In the first stage, an unsupervised sampling strategy is proposed on a large-scale non-precise annotation datasets, forming a multi-stage pre-training process. In the second stage, a fine-tune sampling strategy is proposed on a high-quality dataset containing spontaneous, expressive text and speech pairs. Therefore, the token-wise text-speech association is represented to facilitate alignment between the two modalities. 

\begin{figure}
    \centering
    \includegraphics[width=0.5\linewidth]{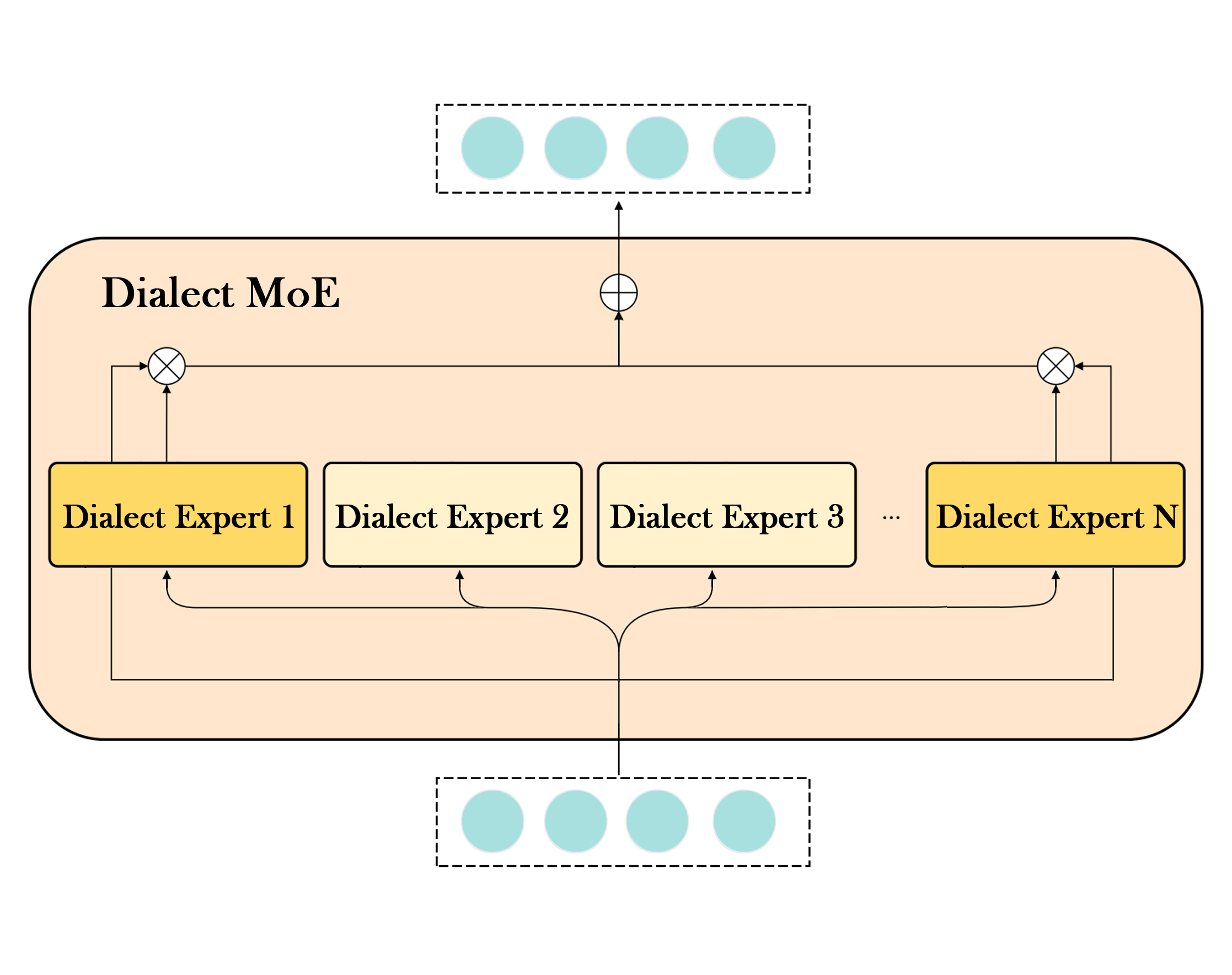}
    \caption{Overall Architecture of Dialectal MoE}
    \label{fig:2}
\end{figure}

\subsection{Multi-Chinese Dialectal Representation Learning}
Dialect-specific TTS models are not satisfactory as dialect-specific data is scarce. Therefore, a single unified TTS model that generalizes well for many dialects is in demand. 

In order to train a unified TTS model for a variety of Chinese dialects, we present a specific token representation together with a multi-stage learning strategy. In the aspect of the token representation, a design of mixture-of-expert architecture (Figure \ref{fig:2}) is proposed to learn the unified representation of multiple Chinese dialects and specific representation of each dialect. Similarly, we propose a cross-attention based mechanism to inject dialect tokens throughout the TTS model layers. 

\subsection{Hierarchical RL-based Posting-training Extension}
Similar to text-based language models, four training stages are built for the representation learning of Bailing-TTS. The first stage is the pre-training stage aiming to maximize scenario and speaker coverage with a robust backbone for general speech representation of the Chinese dialect. This stage is to learn the efficient representation of a large-scale Chinese dialect dataset. The second stage is the fine-tuning stage in order to enhance the basic performance of the dialectal speech. 

Besides, we propose a hierarchical RL learning strategy as the third and the fourth step in order to generate the high-quality speech of multiple Chinese dialects. In the RL hierarchical training strategy, in order to let the high-level policy support exploration for good quality of spontaneous, expressive speech and the primary-level policy support good quality generation of dialectal speech simultaneously. We propose dynamic sub-optimization for the learning of the high-level policy at the base of the primary-level policy in the hierarchical learning strategy.

\section{Experiments and Results}

\subsection{Experimental Settings}
In this section, the details of the training, inference and evaluation for the proposed Bailing-TTS on Chinese dialectal speech synthesis are introduced. 

\subsubsection{Implementation Details}
A custom dataset containing high-quality Mandarin and multiple Chinese dialects data is used for the training, inference and evaluation. This dataset contains $200k$ hours of labeled speech data. The annotation contains fine-grained labels between text and speech modalities. In addition, this large-scale dataset contains high-quality speech-text pairs. Furthermore, this custom Chinese dialect dataset is used for the pre-training stage of the continual semi-supervised learning. Clusters of GPUs (A100) are applied for the training. 

\subsubsection{Evaluation Dataset}
A custom test set is used for the Chinese dialect TTS task. There are $50$ distinct Chinese speakers for each Chinese dialect with $2000$ clips. Following \cite{shen2023naturalspeech}, sentences are randomly chosen from each speaker, and 5-second to 15-second clips are randomly chosen as prompts from the same speaker's speech \cite{shen2023naturalspeech}. Besides, there are $25$ male and $25$ female Chinese speakers for each Chinese dialect. In the dataset, each Chinese dialect contains $50$ common spontaneous words \cite{livingstone2018ryerson}. Besides, $2000$ clips from DiDiSpeech dataset \cite{guo2021didispeech} are used for the evaluation of Mandarin speech.

\subsubsection{Evaluation Metrics}
Both objective metrics and subjective metrics are used for the evaluation. In the aspect of objective metrics: word error rate (WER) and speaker similarity (SIM) is evaluated. In the aspect of subjective metrics, comparative mean option scores (CMOS) and mean option score (MOS) are calculated to evaluate the naturalness and quality, respectively.

\subsection{Experimental Results on TTS of Multiple Chinese Dialects}
In this section, the proposed Bailing-TTS is evaluated in terms of 1)generation quality 2)generation similarity 3)human-level naturalness. 

\subsubsection{Generation Quality}
As shown in Table \ref{tab1}, we find that the proposed Bailing-TTS is close to the ground-truth recording. It demonstrates Bailing-TTS can generate high-quality and natural speech for the testing Chinese dialect data. Besides, the performance verifies the effectiveness of the proposed Bailing-TTS with specific network architecture and the corresponding training strategy. In details, for the evaluation of Mandarin speech, Bailing-TTS obtains $1.86$ for word error rate (WER), $4.32$ for mean option score (MOS). In addition, human speakers obtain $1.35$ for word error rate (WER), $4.32$ for mean option score (MOS). For the evaluation of dialectal speech, Bailing-TTS obtains $6.37$ for word error rate (WER), $4.11$ for mean option score (MOS). Besides, human speakers obtain $4.60$ for word error rate (WER), $4.24$ for mean option score (MOS).



\begin{table}
  \caption{The evaluation results for Bailing-TTS on Chinese speech dataset.}
  \label{tab1}
  \centering
  \begin{tabular}{c|c|c|c|c|c|c}
    \hline
    \multirow{2}{*}{\textbf{Systems}}&$\textbf{WER}\downarrow$&$\textbf{MOS}\uparrow$&$\textbf{CMOS}\uparrow$&$\textbf{WER}\downarrow$&$\textbf{MOS}\uparrow$&$\textbf{CMOS}\uparrow$\\
    \cline{2-4}\cline{5-7}
    &\multicolumn{3}{c}{\textbf{Mandarin}}&\multicolumn{3}{|c}{\textbf{Dialect}}\\
    \hline 
    Human    &$1.35$ &$4.32$ &$-$ &$4.60$ &$4.24$ &$-$\\
    Bailing-TTS  &$1.86$ &$4.21$ &$-0.06$&$6.37$ &$4.11$ &$-0.08$\\
    \bottomrule
  \end{tabular}
\end{table}

\subsubsection{Human-Level Naturalness}
The speech synthesized through Bailing-TTS is also in the comparison with the Ground Truth in Table \ref{tab1} for the evaluation of both Mandarin speech and Chinese dialectal speech. Bailing-TTS achieves $-0.06$ CMOS and $-0.08$ CMOS compared to the Ground Truth, which demonstrates that our method is comparable on human voice quality. 

\subsubsection{Zero-shot learning}
Both objective and subjective evaluation for the zero-shot learning experiments are conducted. In one aspect, clips from custom corpora are used to measure the proposed model's performance on a variety of objective metrics. $2000$ clips from custom dataset are used for the evaluation of Mandarin speech. To be noted, the custom clips contains highly expressive Mandarin speech with spontaneous word.

Besides, both reference utterance and corresponding target utterance are collected from the same speaker for each clip. The proposed Bailing-TTS family is designed to generate target text based on a prompt (the reference speech). Therefore, comparison between synthesized speech and ground truth speech from real human is conducted for both the objective and subjective evaluation.

For the comparison with real human speech, the word error rate (WER) and speaker similarity metrics (SIM) are used for objective evaluation. Particularly, Paraformer-zh \cite{gao2023funasr} is used as the automatic speech recognition (ASR) for Mandarin, and we train a custom automatic speech recognition (ASR) for multiple Chinese dialects. Besides, comparative mean option scores (CMOS) are used for subjective evaluation. Particularly, first, native Chinese dialectal speakers are represented with a reference speech clip of the target speaker. Second, both the synthesized output of the proposed model and the corresponding targeting human speech are used. Third, evaluation is conducted to rate each given clip with higher speaker similarity and expressiveness to the reference clip on a scale between $-3$ to $+3$, where $-3$ and $3$ indicate the least and strongest preference. The results are then collected and averaged over all human evaluators and test sentences. 

The results for both objective and subjective evaluation are reported in Table \ref{tab2}. The results demonstrate that the proposed Bailing-TTS model represents an expected performance in zero-shot learning. In details, for the evaluation of Mandarin speech (Zero-shot), Bailing-TTS obtains $3.48$ for word error rate (WER), $0.72$ for speaker similarity metrics (SIM).

\subsubsection{Fine-tuning learning}
Speaker fine-tuning (SFT) is conducted on the base of pre-trained Bailing-TTS model. For the evaluation of the fine-tuning learning, speech data from $10$ speakers are combined for Mandarin fine-tuning learning dataset. Then, the comparison is made between the fine-tuned model (Bailing-TTS fine-tuned) and the base pre-trained model (Bailing-TTS Zero-shot). For the objective evaluation, the word error rate (WER) and speaker similarity metrics (SIM) are used for the objective evaluation. For the subjective evaluation, comparative mean option scores (CMOS) are used. To be noted, $15$ seconds of speech clip is randomly sampled as the prompt for each speaker. Similarly, speech data from $10$ speakers are combined for Chinese dialectal fine-tuning learning dataset. 

For the evaluation of Mandarin speech, the results of the speaker fine-tuning experiment and the base model are demonstrated in the Table \ref{tab2}. The results demonstrate that the fine-tuned models shows better performance in both objective and subjective metrics. In details, for the evaluation of Mandarin speech (Speaker fine-tuned),  Bailing-TTS obtains $2.98$ for word error rate (WER), $0.77$ for speaker similarity  metrics (SIM) and $+0.18$ for comparative mean option scores (CMOS). For the evaluation of Chinese dialectal speech (Speaker fine-tuned),  Bailing-TTS obtains $7.43$ for word error rate (WER), $0.76$ for speaker similarity metrics (SIM) and $+0.11$ for comparative mean option scores (CMOS). 

\begin{table}[h]
  \caption{The evaluation results of zero-shot learning and speaker fine-tuning for Bailing-TTS on Chinese speech dataset.}
  \label{tab2}
  \centering
  \begin{tabular}{c|c|c|c}
    \hline 
    \multirow{2}{*}{$\textbf{Systems}$}
    &$\textbf{WER}\downarrow$&$\textbf{SIM}\uparrow$&$\textbf{CMOS}\uparrow$\\
    \cline{2-4} 
    &\multicolumn{3}{c}{$\textbf{Mandarin}$}\\
    \hline 
    Bailing-TTS(Zero-shot)&$3.48$ &$0.72$ &$-$\\
    Bailing-TTS(Speaker fine-tuned)  &$2.98$ &$0.77$ &$+0.18$\\
    \hline 
    \textbf{Systems}&\multicolumn{3}{c}{\textbf{Dialect}}\\
    \hline 
    Bailing-TTS(Speaker fine-tuned)  &$7.43$ &$0.76$ &$+0.11$\\
    \bottomrule
  \end{tabular}
\end{table}

\subsubsection{Streaming processing}
There are a range of challenges in the application of TTS systems in the real world. User experiences are not satisfactory due to the latency and the first packet delay. The computation cost of both the time and the memory is still large for the deployment on mobile hardware systems. Therefore, initial work on the deployment is conducted. A variety of techniques \cite{dao2022flashattention,ainslie2023gqa,luo2023latent,lin2306awq} are employed to reduce the inference cost and latency. 

In order to reduce the inference cost and the memory cost of attention layers, first, we apply efficient memory cost methods including grouped-query attention \cite{ainslie2023gqa}, paged attention \cite{kwon2023efficient}, and flash attention \cite{dao2022flashattention,dao2023flashattention}. Second, model quantization method is also applied to further reduce the computation cost \cite{nagel2021white,guo2024decoupleq}. Similarly, the consistency distillation \cite{song2024ella} and a modified flow matching algorithm \cite{esser2024scaling} are used to reduce the computation cost of the diffusion model. Third, to further reduce the computation cost of diffusion architecture, in one aspect, the consistency distillation \cite{song2024ella} is applied to reduce the computation cost of the diffusion model, in the other aspect, the flow matching algorithm \cite{esser2024scaling} is used to reduce the memory cost.

The initial results demonstrate that comparable performance is achieved after a variety of methods are applied on the deployed model. In details, as shown in Table \ref{tab3}, the latency is reduced to $0.13 \times$, the RTF is reduced to $0.46 \times$ and WER and SIM remains the same in the comparison with the offline Bailing-TTS model. 

\begin{table}[h]
  \caption{Comparison between the offline Bailing-TTS model and the online Bailing-TTS model.}
  \label{tab3}
  \centering
  \begin{tabular}{c|c|c|c|c|c}
    \hline \textbf{System}&$\textbf{Latency}\downarrow$&$\textbf{RTF}\downarrow$&$\textbf{WER}\downarrow$&$\textbf{SIM}\uparrow$&$\textbf{CMOS}\uparrow$\\
    \hline
    Offline Bailing-TTS & $1 \times$ &$1 \times$ &$2.75$ &$0.78$ &$-$\\
    \hline
    Online Bailing-TTS  & $0.13 \times$ &$0.46 \times$ &$2.75$ &$0.78$ &$-0.04$\\
    \bottomrule
  \end{tabular}
\end{table}

\section{Applications and Discussion}
We propose a family of Bailing-TTS towards Chinese dialectal speech synthesis with spontaneous words. It has the potential for real-world application. First, the dialectal speech is promising to provide rich experience of chat service in the real-world particularly companion chat service. Second, it's very likely to be beneficial to the facilitation of dialectal culture and cultural applications. 

Initial work on the family of Bailing-TTS is proposed while services such as speech synthesis with emotions, speech with support of other modalities are not well explored. Therefore, the exploration of expressive and emotional generation of Chinese dialectal speech is under-going. We are planing to develop the next version of Bailing-TTS family for the generation of high-quality audio (speech/music) from the input of video and text. Then, the generation of both the high-quality audio together with video will be explored.

\bibliographystyle{plain}
\bibliography{bib}

\begin{thebibliography}{10}

\bibitem{ainslie2023gqa}
Joshua Ainslie, James Lee-Thorp, Michiel de~Jong, Yury Zemlyanskiy, Federico Lebr{\'o}n, and Sumit Sanghai.
\newblock Gqa: Training generalized multi-query transformer models from multi-head checkpoints.
\newblock {\em arXiv preprint arXiv:2305.13245}, 2023.

\bibitem{borsos2023audiolm}
Zal{\'a}n Borsos, Rapha{\"e}l Marinier, Damien Vincent, Eugene Kharitonov, Olivier Pietquin, Matt Sharifi, Dominik Roblek, Olivier Teboul, David Grangier, Marco Tagliasacchi, et~al.
\newblock Audiolm: a language modeling approach to audio generation.
\newblock {\em IEEE/ACM transactions on audio, speech, and language processing}, 31:2523--2533, 2023.

\bibitem{borsos2023soundstorm}
Zal{\'a}n Borsos, Matt Sharifi, Damien Vincent, Eugene Kharitonov, Neil Zeghidour, and Marco Tagliasacchi.
\newblock Soundstorm: Efficient parallel audio generation.
\newblock {\em arXiv preprint arXiv:2305.09636}, 2023.

\bibitem{chen2022wavlm}
Sanyuan Chen, Chengyi Wang, Zhengyang Chen, Yu~Wu, Shujie Liu, Zhuo Chen, Jinyu Li, Naoyuki Kanda, Takuya Yoshioka, Xiong Xiao, et~al.
\newblock Wavlm: Large-scale self-supervised pre-training for full stack speech processing.
\newblock {\em IEEE Journal of Selected Topics in Signal Processing}, 16(6):1505--1518, 2022.

\bibitem{chen2022resgrad}
Zehua Chen, Yihan Wu, Yichong Leng, Jiawei Chen, Haohe Liu, Xu~Tan, Yang Cui, Ke~Wang, Lei He, Sheng Zhao, et~al.
\newblock Resgrad: Residual denoising diffusion probabilistic models for text to speech.
\newblock {\em arXiv preprint arXiv:2212.14518}, 2022.

\bibitem{choi2024dddm}
Ha-Yeong Choi, Sang-Hoon Lee, and Seong-Whan Lee.
\newblock Dddm-vc: Decoupled denoising diffusion models with disentangled representation and prior mixup for verified robust voice conversion.
\newblock In {\em Proceedings of the AAAI Conference on Artificial Intelligence}, volume~38, pages 17862--17870, 2024.

\bibitem{cideron2024musicrl}
Geoffrey Cideron, Sertan Girgin, Mauro Verzetti, Damien Vincent, Matej Kastelic, Zal{\'a}n Borsos, Brian McWilliams, Victor Ungureanu, Olivier Bachem, Olivier Pietquin, et~al.
\newblock Musicrl: Aligning music generation to human preferences.
\newblock {\em arXiv preprint arXiv:2402.04229}, 2024.

\bibitem{dao2023flashattention}
Tri Dao.
\newblock Flashattention-2: Faster attention with better parallelism and work partitioning.
\newblock {\em arXiv preprint arXiv:2307.08691}, 2023.

\bibitem{dao2022flashattention}
Tri Dao, Dan Fu, Stefano Ermon, Atri Rudra, and Christopher R{\'e}.
\newblock Flashattention: Fast and memory-efficient exact attention with io-awareness.
\newblock {\em Advances in Neural Information Processing Systems}, 35:16344--16359, 2022.

\bibitem{deng2023prosody}
Yan Deng, Long Zhou, Yuanhao Yi, Shujie Liu, and Lei He.
\newblock Prosody-aware speecht5 for expressive neural tts.
\newblock In {\em ICASSP 2023-2023 IEEE International Conference on Acoustics, Speech and Signal Processing (ICASSP)}, pages 1--5. IEEE, 2023.

\bibitem{esser2024scaling}
Patrick Esser, Sumith Kulal, Andreas Blattmann, Rahim Entezari, Jonas M{\"u}ller, Harry Saini, Yam Levi, Dominik Lorenz, Axel Sauer, Frederic Boesel, et~al.
\newblock Scaling rectified flow transformers for high-resolution image synthesis.
\newblock In {\em Forty-first International Conference on Machine Learning}, 2024.

\bibitem{gao2023e3}
Yuan Gao, Nobuyuki Morioka, Yu~Zhang, and Nanxin Chen.
\newblock E3 tts: Easy end-to-end diffusion-based text to speech.
\newblock In {\em 2023 IEEE Automatic Speech Recognition and Understanding Workshop (ASRU)}, pages 1--8. IEEE, 2023.

\bibitem{gao2023funasr}
Zhifu Gao, Zerui Li, Jiaming Wang, Haoneng Luo, Xian Shi, Mengzhe Chen, Yabin Li, Lingyun Zuo, Zhihao Du, Zhangyu Xiao, et~al.
\newblock Funasr: A fundamental end-to-end speech recognition toolkit.
\newblock {\em arXiv preprint arXiv:2305.11013}, 2023.

\bibitem{guo2021didispeech}
Tingwei Guo, Cheng Wen, Dongwei Jiang, Ne~Luo, Ruixiong Zhang, Shuaijiang Zhao, Wubo Li, Cheng Gong, Wei Zou, Kun Han, et~al.
\newblock Didispeech: A large scale mandarin speech corpus.
\newblock In {\em ICASSP 2021-2021 IEEE International Conference on Acoustics, Speech and Signal Processing (ICASSP)}, pages 6968--6972. IEEE, 2021.

\bibitem{guo2024decoupleq}
Yi~Guo, Fanliu Kong, Xiaoyang Li, Hui Li, Wei Chen, Xiaogang Tian, Jinping Cai, Yang Zhang, and Shouda Liu.
\newblock decoupleq: Towards 2-bit post-training uniform quantization via decoupling parameters into integer and floating points.
\newblock {\em arXiv preprint arXiv:2404.12759}, 2024.

\bibitem{jiang2023mega}
Ziyue Jiang, Yi~Ren, Zhenhui Ye, Jinglin Liu, Chen Zhang, Qian Yang, Shengpeng Ji, Rongjie Huang, Chunfeng Wang, Xiang Yin, et~al.
\newblock Mega-tts: Zero-shot text-to-speech at scale with intrinsic inductive bias.
\newblock {\em arXiv preprint arXiv:2306.03509}, 2023.

\bibitem{kwon2023efficient}
Woosuk Kwon, Zhuohan Li, Siyuan Zhuang, Ying Sheng, Lianmin Zheng, Cody~Hao Yu, Joseph Gonzalez, Hao Zhang, and Ion Stoica.
\newblock Efficient memory management for large language model serving with pagedattention.
\newblock In {\em Proceedings of the 29th Symposium on Operating Systems Principles}, pages 611--626, 2023.

\bibitem{lajszczak2024base}
Mateusz {\L}ajszczak, Guillermo C{\'a}mbara, Yang Li, Fatih Beyhan, Arent van Korlaar, Fan Yang, Arnaud Joly, {\'A}lvaro Mart{\'\i}n-Cortinas, Ammar Abbas, Adam Michalski, et~al.
\newblock Base tts: Lessons from building a billion-parameter text-to-speech model on 100k hours of data.
\newblock {\em arXiv preprint arXiv:2402.08093}, 2024.

\bibitem{le2024voicebox}
Matthew Le, Apoorv Vyas, Bowen Shi, Brian Karrer, Leda Sari, Rashel Moritz, Mary Williamson, Vimal Manohar, Yossi Adi, Jay Mahadeokar, et~al.
\newblock Voicebox: Text-guided multilingual universal speech generation at scale.
\newblock {\em Advances in neural information processing systems}, 36, 2024.

\bibitem{li2017deep}
Yuxi Li.
\newblock Deep reinforcement learning: An overview.
\newblock {\em arXiv preprint arXiv:1701.07274}, 2017.

\bibitem{lin2306awq}
Ji~Lin, Jiaming Tang, Haotian Tang, Shang Yang, Xingyu Dang, and Song Han.
\newblock Awq: activationaware weight quantization for llm compression and acceleration. corr, abs/2306.00978, 2023. doi: 10.48550.
\newblock {\em arXiv preprint ARXIV.2306.00978}.

\bibitem{livingstone2018ryerson}
Steven~R Livingstone and Frank~A Russo.
\newblock The ryerson audio-visual database of emotional speech and song (ravdess): A dynamic, multimodal set of facial and vocal expressions in north american english.
\newblock {\em PloS one}, 13(5):e0196391, 2018.

\bibitem{lovelace2023simple}
Justin Lovelace, Soham Ray, Kwangyoun Kim, Kilian~Q Weinberger, and Felix Wu.
\newblock Simple-tts: End-to-end text-to-speech synthesis with latent diffusion.
\newblock 2023.

\bibitem{luo2023latent}
Simian Luo, Yiqin Tan, Longbo Huang, Jian Li, and Hang Zhao.
\newblock Latent consistency models: Synthesizing high-resolution images with few-step inference.
\newblock {\em arXiv preprint arXiv:2310.04378}, 2023.

\bibitem{nagel2021white}
Markus Nagel, Marios Fournarakis, Rana~Ali Amjad, Yelysei Bondarenko, Mart Van~Baalen, and Tijmen Blankevoort.
\newblock A white paper on neural network quantization.
\newblock {\em arXiv preprint arXiv:2106.08295}, 2021.

\bibitem{popov2021grad}
Vadim Popov, Ivan Vovk, Vladimir Gogoryan, Tasnima Sadekova, and Mikhail Kudinov.
\newblock Grad-tts: A diffusion probabilistic model for text-to-speech.
\newblock In {\em International Conference on Machine Learning}, pages 8599--8608. PMLR, 2021.

\bibitem{ren2019fastspeech}
Yi~Ren, Yangjun Ruan, Xu~Tan, Tao Qin, Sheng Zhao, Zhou Zhao, and Tie-Yan Liu.
\newblock Fastspeech: Fast, robust and controllable text to speech.
\newblock {\em Advances in neural information processing systems}, 32, 2019.

\bibitem{shen2018natural}
Jonathan Shen, Ruoming Pang, Ron~J Weiss, Mike Schuster, Navdeep Jaitly, Zongheng Yang, Zhifeng Chen, Yu~Zhang, Yuxuan Wang, Rj~Skerrv-Ryan, et~al.
\newblock Natural tts synthesis by conditioning wavenet on mel spectrogram predictions.
\newblock In {\em 2018 IEEE international conference on acoustics, speech and signal processing (ICASSP)}, pages 4779--4783. IEEE, 2018.

\bibitem{shen2023naturalspeech}
Kai Shen, Zeqian Ju, Xu~Tan, Yanqing Liu, Yichong Leng, Lei He, Tao Qin, Sheng Zhao, and Jiang Bian.
\newblock Naturalspeech 2: Latent diffusion models are natural and zero-shot speech and singing synthesizers.
\newblock {\em arXiv preprint arXiv:2304.09116}, 2023.

\bibitem{song2024ella}
Yakun Song, Zhuo Chen, Xiaofei Wang, Ziyang Ma, and Xie Chen.
\newblock Ella-v: Stable neural codec language modeling with alignment-guided sequence reordering.
\newblock {\em arXiv preprint arXiv:2401.07333}, 2024.

\bibitem{tan2024naturalspeech}
Xu~Tan, Jiawei Chen, Haohe Liu, Jian Cong, Chen Zhang, Yanqing Liu, Xi~Wang, Yichong Leng, Yuanhao Yi, Lei He, et~al.
\newblock Naturalspeech: End-to-end text-to-speech synthesis with human-level quality.
\newblock {\em IEEE Transactions on Pattern Analysis and Machine Intelligence}, 2024.

\bibitem{wang2023neural}
Chengyi Wang, Sanyuan Chen, Yu~Wu, Ziqiang Zhang, Long Zhou, Shujie Liu, Zhuo Chen, Yanqing Liu, Huaming Wang, Jinyu Li, et~al.
\newblock Neural codec language models are zero-shot text to speech synthesizers.
\newblock {\em arXiv preprint arXiv:2301.02111}, 2023.

\bibitem{wang2024speechx}
Xiaofei Wang, Manthan Thakker, Zhuo Chen, Naoyuki Kanda, Sefik~Emre Eskimez, Sanyuan Chen, Min Tang, Shujie Liu, Jinyu Li, and Takuya Yoshioka.
\newblock Speechx: Neural codec language model as a versatile speech transformer.
\newblock {\em IEEE/ACM Transactions on Audio, Speech, and Language Processing}, 2024.

\bibitem{wang2017tacotron}
Yuxuan Wang, RJ~Skerry-Ryan, Daisy Stanton, Yonghui Wu, Ron~J Weiss, Navdeep Jaitly, Zongheng Yang, Ying Xiao, Zhifeng Chen, Samy Bengio, et~al.
\newblock Tacotron: Towards end-to-end speech synthesis.
\newblock {\em arXiv preprint arXiv:1703.10135}, 2017.

\bibitem{wang2023lm}
Zhichao Wang, Yuanzhe Chen, Lei Xie, Qiao Tian, and Yuping Wang.
\newblock Lm-vc: Zero-shot voice conversion via speech generation based on language models.
\newblock {\em IEEE Signal Processing Letters}, 2023.

\bibitem{yu2023language}
Lijun Yu, Jos{\'e} Lezama, Nitesh~B Gundavarapu, Luca Versari, Kihyuk Sohn, David Minnen, Yong Cheng, Agrim Gupta, Xiuye Gu, Alexander~G Hauptmann, et~al.
\newblock Language model beats diffusion--tokenizer is key to visual generation.
\newblock {\em arXiv preprint arXiv:2310.05737}, 2023.

\bibitem{zeghidour2021soundstream}
Neil Zeghidour, Alejandro Luebs, Ahmed Omran, Jan Skoglund, and Marco Tagliasacchi.
\newblock Soundstream: An end-to-end neural audio codec.
\newblock {\em IEEE/ACM Transactions on Audio, Speech, and Language Processing}, 30:495--507, 2021.

\bibitem{zhang2024speechalign}
Dong Zhang, Zhaowei Li, Shimin Li, Xin Zhang, Pengyu Wang, Yaqian Zhou, and Xipeng Qiu.
\newblock Speechalign: Aligning speech generation to human preferences.
\newblock {\em arXiv preprint arXiv:2404.05600}, 2024.

\bibitem{zhang2023speak}
Ziqiang Zhang, Long Zhou, Chengyi Wang, Sanyuan Chen, Yu~Wu, Shujie Liu, Zhuo Chen, Yanqing Liu, Huaming Wang, Jinyu Li, et~al.
\newblock Speak foreign languages with your own voice: Cross-lingual neural codec language modeling.
\newblock {\em arXiv preprint arXiv:2303.03926}, 2023.

\end{thebibliography}
\end{document}